\documentclass{article}

\usepackage{microtype}
\usepackage{graphicx}
\usepackage{subcaption}
\usepackage{booktabs} 
\usepackage[most]{tcolorbox}
\tcbuselibrary{breakable}

\usepackage{hyperref}

\usepackage[accepted]{icml2026}

\usepackage{amsmath}
\usepackage{amssymb}
\usepackage{mathtools}
\usepackage{amsthm}

\usepackage[capitalize,noabbrev]{cleveref}

\theoremstyle{plain}

\theoremstyle{definition}

\theoremstyle{remark}

\usepackage[textsize=tiny]{todonotes}

\icmltitlerunning{ToolFailBench: Diagnosing Tool-Use Failures in LLM Agents}

\newtcblisting{promptlisting}[1][]{
  breakable,
  listing only,
  colback=gray!3,
  colframe=gray!35,
  boxrule=0.5pt,
  arc=2pt,
  left=6pt,
  right=6pt,
  top=6pt,
  bottom=6pt,
  listing options={
    basicstyle=\ttfamily\scriptsize,
    breaklines=true,
    breakatwhitespace=false,
    columns=fullflexible,
    keepspaces=true,
    showstringspaces=false
  },
  #1
}

\begin{document}

\twocolumn[
  \icmltitle{ToolFailBench: Diagnosing Tool-Use Failures in LLM Agents}



\begin{icmlauthorlist}
\icmlauthor{Harsh Soni}{berkeley}
\end{icmlauthorlist}

\icmlaffiliation{berkeley}{UC Berkeley}
\icmlcorrespondingauthor{Harsh Soni}{harsh.soni@berkeley.edu}

  \icmlkeywords{Machine Learning, ICML}

  \vskip 0.3in
]



\printAffiliationsAndNotice{}  

\begin{abstract}
Tool calling is central to modern language model agents, but aggregate benchmark scores often hide where tool use fails. A model that never calls a needed tool and a model that calls the tool but ignores the result can look similar under final task accuracy. We introduce \emph{ToolFailBench}, a diagnostic benchmark for measuring tool-use failures across 1{,}000 tasks in finance, medicine, law, cybersecurity, and real estate. Tool-required tasks return values the model wouldn't guess, forcing it to trust the tool while control tasks attach the same tools but should be answered directly. We label each trace with Tool-Skip, Result-Ignore, Output-Fabrication, and Unnecessary-Tool-Use, using a rule classifier and two LLM judges aggregated by majority vote. Across 19 headline models, the best reaches 86.33\% Clean Tool-Use Rate, showing that faithful tool use is not saturated. More importantly, models with similar aggregate scores fail in different ways: most stay disciplined on no-tool controls, while Llama-3.1 models show an Always-Call pattern, and at the same parameter scale Llama-3.1-70B and Qwen2.5-72B differ by 89 percentage points on control-task accuracy. Tool-use evaluation should measure not only whether agents call tools, but whether they use tool outputs correctly and avoid tools when none is needed. Code:\,\href{https://github.com/SoHarshh/ToolFailBench}{github.com/SoHarshh/ToolFailBench}
 
\end{abstract}

\section{Introduction}
\label{sec:intro}

Tool calling is now a core part of language model agents. Modern agents are expected to call external functions, APIs, search tools, databases, and domain-specific services, then use the returned information in their final answer~\citep{schick2023toolformer,patil2025bfcl}. This matters in settings such as customer support~\citep{yao2024taubench}, finance~\citep{bigeard2025fab}, healthcare~\citep{jiang2025medagentbench}, software engineering~\citep{jimenez2024swebench}, and long-horizon multi-application workflows~\citep{li2025toolathlon}. In these settings, reliability depends not only on whether a model calls a tool, but also on whether it uses the tool result correctly.

Existing benchmarks measure important parts of tool use, but they often collapse different failures into one score. BFCL evaluates function-call correctness~\citep{patil2025bfcl}; $\tau$-bench evaluates whether an agent reaches the right end state in simulated customer-service tasks~\citep{yao2024taubench}; ToolLLM measures tool-use success across real-world APIs~\citep{qin2024toolllm}; and Toolathlon evaluates long-horizon task execution across many software applications~\citep{li2025toolathlon}. These benchmarks are useful, but a single pass rate or accuracy number may hide "why" a model failed. A model that never calls a needed tool, a model that calls the tool but ignores the result, and a model that calls the tool but invents extra information can all look similar under aggregate evaluation, even though they fail in different ways.

We introduce \textbf{ToolFailBench}, a diagnostic benchmark for measuring where tool-use failures occur. ToolFailBench contains 1{,}000 single-turn tasks across five professional domains: finance, medicine, law, cybersecurity, and real estate. Tool-required tasks are built as \emph{parametric traps}: the mock tool return contradicts a likely memorized prior, so a model must use the returned value instead of falling back on memory. Control tasks test the opposite behavior, asking whether a model can answer directly without calling a tool unnecessarily.

ToolFailBench assigns each response a failure-mode label rather than only a final success score. For tool-required tasks, we separate \emph{Tool-Skip}, \emph{Result-Ignore}, \emph{Output-Fabrication}, and correct behavior. For control tasks, we separate \emph{Unnecessary-Tool-Use}, wrong direct answers, and correct direct answers. We label each trace with a deterministic rule classifier and two independent LLM judges, then aggregate the labels by majority vote. This gives us a reproducible rule-only view and a semantic ensemble view of the same model behavior.

Across 19 headline models with valid traces, ToolFailBench is not saturated: the best model reaches 86.33\% Clean Tool-Use Rate. More importantly, similar aggregate scores hide different failure profiles. Most models form a disciplined low-UTR cluster, while Llama-3.1 models show an Always-Call pattern. At the same parameter scale, Llama-3.1-70B and Qwen2.5-72B produce opposite behavior on identical inputs, differing by 89 percentage points on control-task accuracy. These results suggest that tool discipline is not explained by scale alone, but depends strongly on model family and training behavior.

\noindent\textbf{Contributions.}
We make three contributions. First, we introduce ToolFailBench, a 1{,}000-task diagnostic benchmark across five professional domains for testing whether agents use tool outputs correctly. Second, we define a failure-mode taxonomy that separates skipped tools, ignored tool results, fabricated outputs, and unnecessary tool use. Third, we evaluate 19 headline models and show that aggregate scores hide distinct failure profiles, including a same-scale Llama-vs-Qwen contrast with an 89-point gap in control-task accuracy.

\section{Related Work}
\label{sec:related}

\paragraph{Tool-use and function-calling benchmarks.}
Early tool-use work studied how language models can learn to call external tools and APIs, including through self-supervised tool-use traces and API-documentation retrieval~\citep{schick2023toolformer,patil2023gorilla}. Subsequent benchmarks evaluate tool use more directly. API-Bank studies tool-augmented dialogue with runnable APIs~\citep{li2023apibank}; ToolLLM scales tool-use data and evaluation to thousands of real-world APIs~\citep{qin2024toolllm}; and BFCL evaluates function-calling correctness, including whether models produce valid and executable calls~\citep{patil2025bfcl}. These benchmarks are central for measuring tool selection and call validity. ToolFailBench focuses on a different part of the loop: after a tool return is available, does the model actually use it in the final answer?

\paragraph{Agentic task-completion benchmarks.}
Another line of work evaluates agents in richer environments where tool use is embedded inside a larger task. $\tau$-bench measures task completion in simulated customer-service interactions~\citep{yao2024taubench}; SWE-bench evaluates software issue resolution~\citep{jimenez2024swebench}; and domain benchmarks study agents in settings such as medicine and finance~\citep{jiang2025medagentbench,bigeard2025fab}. ToolSandbox, AppWorld, and Toolathlon further test stateful, interactive, and long-horizon tool use across applications~\citep{lu2024toolsandbox,trivedi2024appworld,li2025toolathlon}. These environments are valuable because they test realistic end-to-end behavior, but final success can mix planning, retrieval, tool choice, execution, and answer integration. ToolFailBench is narrower by design: it isolates whether the returned tool evidence is used faithfully.

\paragraph{Diagnostic evaluation of tool-use failures.}
Recent work has begun to analyze tool-use failures beyond aggregate success rates. T-Eval decomposes tool utilization into intermediate capabilities such as instruction following, planning, retrieval, reasoning, and review~\citep{chen2023teval}. StableToolBench studies instability in tool-use evaluation and proposes more stable protocols~\citep{guo2024stabletoolbench}. SpecTool characterizes error patterns in tool-use outputs~\citep{kokane2024spectool}, ToolBeHonest studies hallucination and honesty failures in tool-augmented settings~\citep{zhang2024toolbehonest}, and SMART studies tool overuse, where agents call tools for queries answerable without external tools~\citep{qian2025smart}. ToolFailBench is closest to this diagnostic line of work, but targets a specific missing behavior: post-call faithfulness under controlled tool returns. By separating Tool-Skip, Result-Ignore, Output-Fabrication, and Unnecessary-Tool-Use, ToolFailBench complements prior benchmarks by showing not only whether a model succeeds, but where the tool-use loop breaks.

\section{ToolFailBench}
\label{sec:toolfailbench}

ToolFailBench is a diagnostic benchmark for testing whether language model agents use tool results when writing their final answers. The benchmark contains 1{,}000 tasks across five professional domains and tests both tool-required queries and control queries where no tool is needed. For each response, ToolFailBench checks the model's tool decision, returned evidence, and final answer, then assigns a failure-mode label. To make labeling reliable, we combine a deterministic rule classifier with two independent LLM judges and report majority-vote ensemble results.

\begin{figure*}[t]
\centering
\includegraphics[width=1\textwidth]{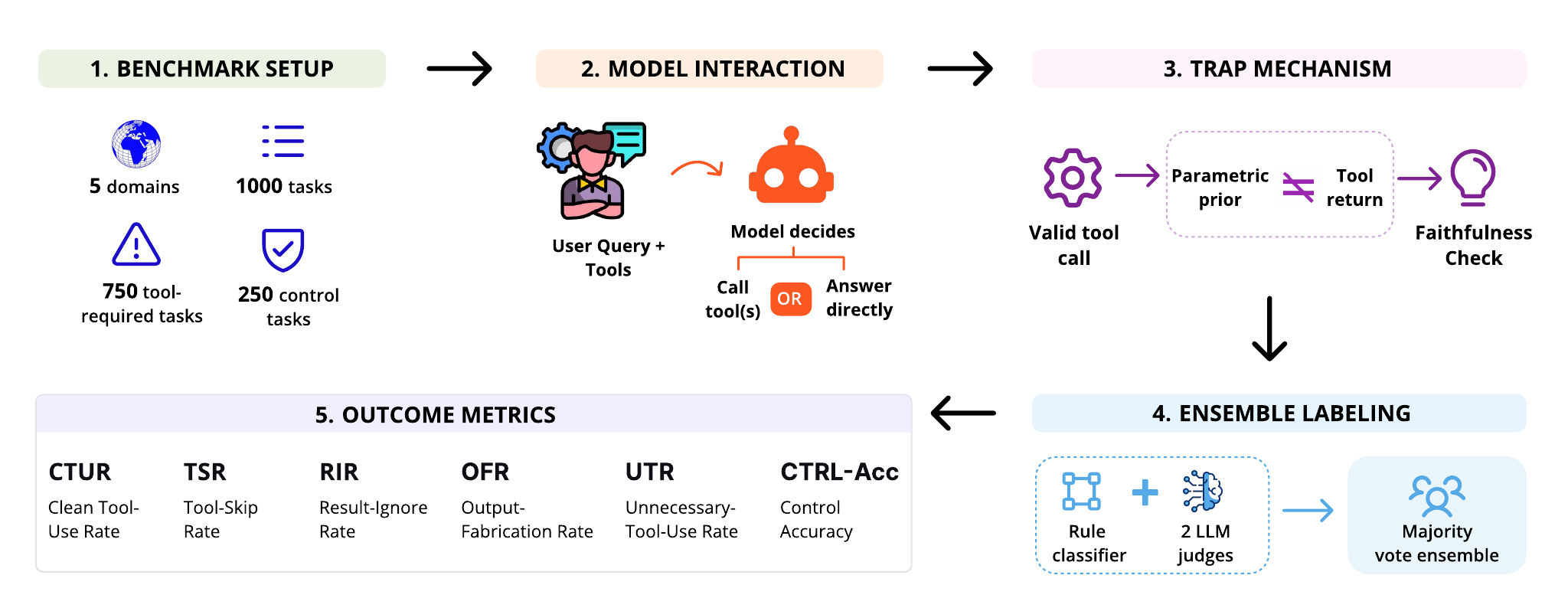}
\vspace{-6pt}
\caption{ToolFailBench overview. The benchmark contains 1{,}000 single-turn tasks across five professional domains, with 750 tool-required parametric traps and 250 control tasks. In each trap, the mock tool return contradicts a plausible parametric prior, so the model's answer reveals whether it follows the tool return or its prior. Each trace is labeled by a rule classifier and two LLM judges via majority vote, producing the failure-mode metrics in Table~\ref{tab:metrics}.}
\label{fig:workflow}
\vspace{-8pt}
\end{figure*}

\subsection{Benchmark Design}
\label{sec:design}

ToolFailBench follows a controlled tool-use setting: models receive a user query, access to domain tools, and must decide whether to call a tool before producing a final answer~\citep{patil2025bfcl,yao2024taubench,li2025toolathlon}. Unlike benchmarks that primarily score tool selection or end-task completion, ToolFailBench is designed to show where the tool-use loop breaks: whether the model skips a needed tool, ignores the tool result, fabricates unsupported information, or calls a tool when none is needed.

The benchmark contains 1{,}000 tasks across five professional domains: finance, medicine, law, cybersecurity, and real estate. Each domain contains 200 tasks, split into 150 tool-required tasks and 50 control tasks where no tool is needed. Tool-required tasks are constructed as \emph{parametric traps}: the mock tool return intentionally differs from a plausible prior value the model may have stored in parametric memory. A faithful model should use the returned value, while an unfaithful answer may replace it with the prior value or another unsupported value. We use ``memorized prior'' as a task-design term rather than a claim that the benchmark directly identifies the model's internal source of the value; additional construction details are given in Appendix~\ref{app:trap-values}.

Each tool-required task includes a specific entity, an expected tool, an expected answer, and the values needed for scoring. For trap tasks, we store both the tool-returned value and the likely memorized value. For control tasks, we store the expected direct answer and check whether the model avoids unnecessary tool use.

\subsection{Failure Mode Taxonomy}
\label{sec:taxonomy}

Most function-calling and agent benchmarks report tool-call accuracy, final task success, or LLM-judged pass rates~\citep{patil2025bfcl,yao2024taubench,qin2024toolllm,guo2024stabletoolbench}. These scores are useful, but they do not show why a tool-using model failed. Following recent diagnostic work on tool-use errors~\citep{kokane2024spectool}, ToolFailBench assigns each response one label that separates tool skipping, result ignoring, fabrication, and unnecessary tool use.

For tool-required tasks, a response is \emph{Correct} if the model calls the expected tool and uses the returned data in its final answer. We define three failure modes: \textbf{Tool-Skip (TS)}, where the model does not produce a valid executed tool call; \textbf{Result-Ignore (RI)}, where the model calls the tool but does not use the returned data; and \textbf{Output-Fabrication (OF)}, where the model calls the tool but adds invented structured information not present in the return.

For control tasks, where no tool is needed, a response is \emph{Correct} if the model answers directly without calling a tool. We label \textbf{Unnecessary-Tool-Use (UTU)} when the model calls a tool anyway, and \textbf{Wrong-Answer} when the model avoids tool use but answers incorrectly. Each response receives exactly one label within its task type, allowing ToolFailBench to distinguish models that achieve similar aggregate scores but fail for different reasons.

\subsection{System Prompt and Output Format}
\label{sec:prompt}

Each domain uses one shared system prompt for all models, with no model-specific tuning and no in-context examples containing task answers. The prompt tells the model when to call a tool, when to answer directly, and how to format the final response. We use a uniform answer format so outputs can be compared across models and domains, and so the rule classifier has a stable surface for checking whether the final answer uses the returned value. Full prompts and output templates are provided in Appendix~\ref{app:prompts}.

\subsection{Metrics and Detection}
\label{sec:metrics}

Each model produces one trace per task containing the user query, executed tool calls, mock tool returns, and final answer. We score each trace with a deterministic rule classifier and two independent LLM judges. The final label is the majority vote over the three raters, with ties resolved by the rule classifier.

\paragraph{Metrics.}
We report six metrics over three task sets. \emph{Required} contains the 750 tasks where a tool should be called. \emph{Called} is the subset of required tasks where the model actually made an executed tool call. \emph{Control} contains the 250 tasks where the model should answer without using a tool.

This separation matters because a model can fail at different points in the tool-use loop. TSR and CTUR measure whether the model calls a tool when needed. RIR and OFR measure post-call failures after a tool has already been called. UTR and CTRL-Acc measure whether the model avoids tools when no tool is needed.

\begin{table}[H]
\centering
\begin{minipage}{1\columnwidth}
\centering
\small
\setlength{\tabcolsep}{4pt}
\begin{tabular*}{\linewidth}{@{\extracolsep{\fill}}lll@{}}
\toprule
\textbf{Metric} & \textbf{Name} & \textbf{Set} \\
\midrule
TSR & Tool-Skip Rate & Required \\
CTUR & Clean Tool-Use Rate & Required \\
RIR & Result-Ignore Rate & Required \\
OFR & Output-Fabrication Rate & Required \\
UTR & Unnecessary-Tool-Use Rate & Control \\
CTRL-Acc & Control Accuracy & Control \\
\bottomrule
\end{tabular*}
\vspace{2pt}
\caption{ToolFailBench metric definitions. Required = 750 tool-needed tasks; Control = 250 no-tool tasks. TSR + CTUR + RIR + OFR sum to 100\% on Required, so leaderboard rows are verifiable by addition.}
\label{tab:metrics}
\end{minipage}
\vspace{-16pt}
\end{table}

\paragraph{Detection.}
The rule classifier assigns labels from the trace without any LLM calls. Tool-Skip and Unnecessary-Tool-Use are determined by whether an executed tool call appears in the trace. Result-Ignore is detected when the final answer omits the expected tool-returned value, and Output-Fabrication is detected when the answer contains invented structured information not supported by the mock return. Because rule-based checks can miss faithful paraphrases, we also score each trace with two LLM judges using the same failure-mode rubric~\citep{zheng2023judging}. Full judge and rule-classifier details are provided in Appendix~\ref{app:judge-rubrics}.

\section{Experimental Setup}
\subsection{Models}
\label{sec:models}

We evaluate instruction-following models from closed-API providers, open-weight model families, and reasoning-oriented models. The headline comparison uses 19 models that completed the full ToolFailBench suite with valid traces. The model set spans OpenAI, Anthropic, xAI, DeepSeek, Qwen, Llama, Gemma, GLM, Mistral, and QwQ models.

We ran 22 models in total. Nineteen are included in the headline leaderboard, two are reported as non-headline tool-skip cases, and one is excluded because a chat-template/detokenizer issue produced invalid raw-token output. Full model identifiers and run statuses are listed in Appendix~\ref{app:model-identifiers}.

\subsection{Inference Protocol}
\label{sec:inference}

All models use the same task set, prompts, tool schemas, and decoding settings. We use \texttt{temperature=0} and \texttt{max\_tokens=1024} for every run to keep traces reproducible and avoid mixing tool-use failures with sampling variance~\citep{renze2024temperature,song2024greedy}. We set \texttt{seed=42} where supported; Anthropic does not expose a seed parameter, so Claude runs are deterministic through \texttt{temperature=0} only.

For each task, the model first receives the user query with the available tools attached and \texttt{tool\_choice=auto}. If the model emits a valid tool call, we execute the corresponding mock tool and inject the controlled return as a tool message. The model then produces a final natural-language answer with \texttt{tool\_choice=none}. This two-step protocol separates the tool-decision step from the answer-writing step, allowing us to distinguish skipped tools, ignored tool returns, and fabricated outputs.

Closed-API models are queried through their provider APIs. Open-weight and reasoning models are served through OpenAI-compatible vLLM endpoints~\citep{kwon2023vllm}. Each model is run on the full 1{,}000-task suite. Additional serving details are provided in Appendix~\ref{app:inference-settings}.

\subsection{Judge Configuration}
\label{sec:judge-config}

We use two LLM judges because exact rule checks are reproducible but brittle: they can miss faithful paraphrases when the model uses the correct tool value without copying the expected string. To reduce prompt-specific blind spots, we use two judges with different prompt framings. Qwen3.5-397B-A17B-FP8 uses a decision-tree prompt, while GLM-4.7-FP8 uses an evidence-attribution prompt. Both judges receive the same trace and apply the failure-mode taxonomy from Section~\ref{sec:taxonomy}; the final ensemble label is the majority vote over the rule classifier and the two judges, with rare three-way ties resolved by the rule classifier.

Because several evaluated models are from the Qwen family, we also check for same-family judge bias. A same-family preference would predict the Qwen judge to score Qwen-family models more leniently than the GLM judge. Instead, the Qwen judge is stricter for both Qwen-family and non-Qwen models, with nearly identical mean gaps: -5.99 percentage points for Qwen-family models and -6.16 percentage points for non-Qwen models. We report rule-only metrics, ensemble metrics, and inter-rater agreement in Section~\ref{sec:agreement}, with the same-family check in Appendix~\ref{app:same-family-judge-check}.

\section{Results and Analysis}
\label{sec:results}

We report main results on the 19 models that completed the full ToolFailBench suite with valid traces. Unless stated otherwise, results use the ensemble label from the rule classifier and two LLM judges. Rule-only numbers are reported where useful because they are fully deterministic from the released traces. Three additional runs are excluded from the headline leaderboard and documented in Appendix~\ref{app:nonheadline}: two tool-style answer cases without clean executed tool calls and one invalid raw-token run.

\begin{table*}[t]
\centering
\small
\setlength{\tabcolsep}{3pt}
\caption{Main leaderboard (19 models, sorted by ensemble CTUR). Grok-4.3 attains the highest CTUR at 86.33\%, but the strongest models are not those that call tools most often --- they call tools when needed and use the returned values faithfully. Llama-3.1 models stand out as Always-Call outliers (UTR $\geq$ 77\%). Higher is better for CTUR and CTRL-Acc; lower is better for TSR, RIR, OFR, UTR. ``Ens.'' = majority vote across the rule classifier and two LLM judges; Rule CTUR uses the rule classifier alone. Metrics defined in Table~\ref{tab:metrics}.}
\label{tab:leaderboard}
\begin{tabular*}{\textwidth}{@{\extracolsep{\fill}}r l r r r r r r r@{}}
\toprule
\# & Model & Ens. CTUR & Rule CTUR & Ens. TSR & Ens. RIR & Ens. OFR & Ens. UTR & Ens. CTRL-Acc \\
\midrule
1  & {Grok-4.3}                 & {86.33} & 80.67 & 11.80 & 1.74  & 0.13 & 0.81  & 97.18 \\
2  & Grok-4-1-Fast-Reasoning           & 84.11 & 80.67 & 14.15 & 1.60  & 0.13 & 0.40  & 99.60 \\
3  & Qwen2.5-32B-Instruct              & 82.68 & 78.27 & 12.08 & 4.43  & 0.81 & 0.80  & 95.58 \\
4  & Qwen3.6-27B                       & 79.33 & 76.67 & 19.87 & 0.67  & 0.13 & 0.00  & 98.79 \\
5  & Claude-Sonnet-4.5                   & 79.28 & 73.07 & 15.64 & 4.41  & 0.67 & 0.00  & 100.00 \\
6  & GPT-5.4-Mini                      & 79.14 & 72.27 & 14.17 & 5.48  & 1.20 & 0.00  & 97.20 \\
7  & QwQ-32B                           & 79.04 & 75.47 & 16.02 & 3.07  & 1.87 & 1.61  & 95.98 \\
8  & Qwen2.5-72B-Instruct              & 79.00 & 75.20 & 18.57 & 2.02  & 0.40 & 0.00  & 98.00 \\
9  & Qwen3.6-35B-A3B                   & 78.47 & 75.47 & 18.98 & 2.15  & 0.40 & 0.00  & 99.60 \\
10 & Gemma-4-31B                       & 78.12 & 70.80 & 19.06 & 2.68  & 0.13 & 0.00  & 98.39 \\
11 & Qwen3.5-27B                       & 77.38 & 75.47 & 21.55 & 0.67  & 0.40 & 0.00  & 99.60 \\
12 & Claude-Haiku-4.5                  & 76.47 & 68.93 & 18.85 & 2.81  & 1.87 & 0.00  & 100.00 \\
13 & DeepSeek-V4-Flash                 & 75.84 & 71.33 & 17.45 & 4.30  & 2.42 & 1.20  & 98.39 \\
14 & Gemma-4-27B-A4B                   & 73.49 & 68.40 & 24.10 & 1.74  & 0.67 & 0.00  & 99.20 \\
15 & GLM-4.7-Flash                     & 71.49 & 68.80 & 24.90 & 3.21  & 0.40 & 0.00  & 99.19 \\
16 & Qwen3.5-9B                        & 70.03 & 70.40 & 26.48 & 2.82  & 0.67 & 0.00  & 99.60 \\
17 & Qwen2.5-7B-Instruct               & 65.28 & 60.80 & 28.53 & 5.11  & 1.08 & 0.00  & 95.16 \\
18 & Llama-3.1-70B                     & 62.58 & 59.73 & 24.23 & 11.17 & 2.02 & 77.73 & 8.91 \\
19 & Llama-3.1-8B                      & 47.32 & 47.60 & 20.64 & 30.43 & 1.61 & 98.39 & 0.00 \\
\bottomrule
\end{tabular*}
\end{table*}

\subsection{Main Leaderboard}
\label{sec:headline}

Table~\ref{tab:leaderboard} gives the main leaderboard over the 19 headline models, sorted by ensemble Clean Tool-Use Rate (CTUR). Grok-4.3 attains the highest CTUR at 86.33\%, followed by Grok-4-1-Fast-Reasoning at 84.11\% and Qwen2.5-32B-Instruct at 82.68\%. These top scores are statistically close: their 95\% Wilson confidence intervals overlap, and Grok-4.3 is not significantly above Grok-4-1-Fast-Reasoning ($p=0.23$) and only marginally above Qwen2.5-32B-Instruct ($p=0.05$). We therefore treat the strongest models as a leading cluster rather than a strict ranking. ToolFailBench is not saturated: even the strongest model still fails to cleanly use tools on a meaningful fraction of tool-required tasks.

High CTUR does not simply mean calling tools more often. The strongest models call tools when needed and use the returned evidence correctly, while keeping Result-Ignore and Output-Fabrication rates low.

\subsection{Failure Profiles Reveal Different Model Behaviors}
\label{sec:archetypes}

\begin{figure}[!ht]
\centering
\includegraphics[width=0.9\columnwidth]{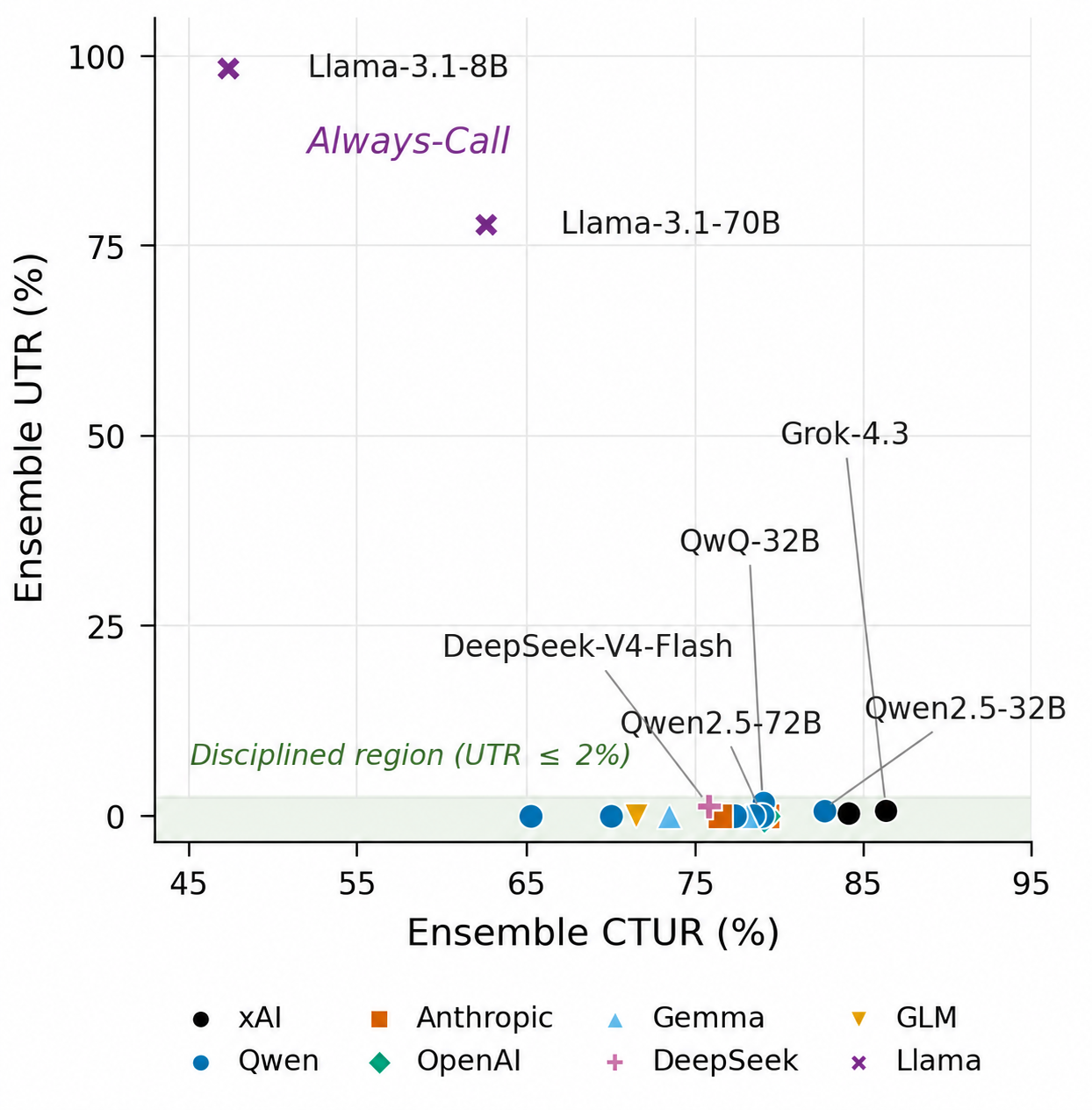}
\vspace{-5pt}
\caption{CTUR--UTR failure profiles. Most models remain disciplined, while Llama-3.1 models form Always-Call outliers.}
\label{fig:archetype}
\vspace{-18pt}
\end{figure}

A leaderboard score can show which model performs better, but not how the model fails. In ToolFailBench, models with similar Clean Tool-Use Rate (CTUR) can still fail in different ways: they may skip a needed tool, call the tool but ignore the returned evidence, or call tools even when no tool is needed. Figure~\ref{fig:archetype} makes this visible by comparing CTUR with Unnecessary-Tool-Use Rate (UTR).

The main pattern is a large \emph{Disciplined} cluster. Most models achieve moderate to high CTUR while keeping UTR near zero, meaning they rarely call tools on control questions where no tool is needed. This cluster includes the closed-API models, the Qwen models, Gemma, GLM-4.7-Flash, QwQ-32B, and DeepSeek-V4-Flash. Across these models, UTR stays at or below 1.61\%, and CTRL-Acc stays above 95\%.

The Llama-3.1 models form a different regime. Llama-3.1-70B reaches 62.58\% CTUR, but its UTR is 77.73\% and its CTRL-Acc is only 8.91\%. Llama-3.1-8B is more extreme, with 47.32\% CTUR, 98.39\% UTR, and 0.00\% CTRL-Acc. These models often call tools even when the question should be answered directly, forming an \emph{Always-Call} pattern.

This separation is the point of ToolFailBench. Models with similar aggregate tool-use scores can fail for different reasons, and those differences are not visible from a leaderboard alone.

\subsection{Same Scale Does Not Mean Same Tool Discipline}
\label{sec:family-scale}

The Always-Call pattern is not only a small-model issue. Scaling Llama-3.1 from 8B to 70B improves CTUR and reduces Result-Ignore Rate, but it does not fix control-task behavior: Llama-3.1-70B still calls tools on 77.73\% of no-tool control tasks. The same-scale comparison with Qwen2.5 is sharper: Qwen2.5-72B reaches higher CTUR, much lower RIR, zero UTR, and far higher control accuracy. The 89-point control-accuracy gap between Llama-3.1-70B and Qwen2.5-72B is highly significant (two-proportion $z=-19.9$, $p<10^{-80}$).

This suggests that tool discipline is not explained by parameter count alone. Larger models can improve clean tool use, but avoiding unnecessary tool calls appears to depend on model family and training behavior. We treat the Llama-3.1 Always-Call pattern as family-specific.

\begin{table}[!ht]
\centering
\begin{minipage}{1\columnwidth}
\centering
\small
\setlength{\tabcolsep}{3pt}
\begin{tabular*}{\linewidth}{@{\extracolsep{\fill}}lrrrr@{}}
\toprule
\textbf{Model} & \textbf{CTUR} & \textbf{RIR} & \textbf{UTR} & \textbf{CTRL-Acc} \\
\midrule
Llama-3.1-8B          & 47.32 & 30.43 & 98.39 & 0.00 \\
Llama-3.1-70B         & 62.58 & 11.17 & 77.73 & 8.91 \\
Qwen2.5-7B-Instruct   & 65.28 & 5.11  & 0.00  & 95.16 \\
Qwen2.5-72B-Instruct  & 79.00 & 2.02  & 0.00  & 98.00 \\
\bottomrule
\end{tabular*}
\vspace{2pt}
\caption{Scale comparison between Llama-3.1 and Qwen2.5 models. All values are percentages.}
\label{tab:scale-comparison}
\end{minipage}
\vspace{-6pt}
\end{table}


\subsection{Labeling Robustness}
\label{sec:agreement}

We use ensemble labels for the headline results because exact rule checks are reproducible but brittle. The rule classifier can mark a faithful answer as a failure when the model uses the correct tool value but does not copy the expected string exactly. The two LLM judges reduce this surface-form brittleness by applying the same failure-mode rubric semantically, while the rule-only scores remain useful as a deterministic audit trail.

\begin{table}[H]
\centering
\begin{minipage}{1\columnwidth}
\centering
\small
\setlength{\tabcolsep}{7pt}
\begin{tabular*}{\linewidth}{@{\extracolsep{\fill}}lr@{}}
\toprule
\textbf{Rater pair} & \textbf{Mean $\kappa$} \\
\midrule
Rule vs.\ Qwen judge      & 0.649 \\
Rule vs.\ GLM judge       & 0.664 \\
Qwen judge vs.\ GLM judge & 0.773 \\
Three-way Fleiss' $\kappa$ & 0.693 \\
\bottomrule
\end{tabular*}
\vspace{2pt}
\caption{Agreement between the deterministic rule classifier and the two LLM judges. Higher $\kappa$ indicates more consistent failure-mode labels across raters.}
\label{tab:kappa}
\end{minipage}
\vspace{-6pt}
\end{table}

The judge-judge agreement is higher than either rule-judge agreement, which is expected if the judges recognize faithful paraphrases that surface rules may miss. Across the 19 headline models, mean ensemble CTUR is 75.02\%, while mean rule CTUR is 71.05\%. The ensemble label therefore makes scoring less brittle without changing the main leaderboard story.

The rule classifier acts as the ensemble tiebreaker only on genuine three-way disagreements, which occur in just 1.5\% of judgments (285 of 18{,}900). By contrast, the two LLM judges jointly overrule the rule label in 8.4\% of judgments. Thus, the ensemble revises deterministic labels more often than it falls back to them, reducing dependence on surface-level rule decisions.

Agreement is also stable across strata: Appendix~\ref{app:kappa} breaks down inter-rater agreement by domain, failure mode, and model tier, showing that lower $\kappa$ values for rare labels mainly reflect low prevalence rather than broad rater disagreement.

\subsection{Domain Effects}
\label{sec:domain}

Tool-result faithfulness also varies by domain. To isolate post-call behavior, we compare rule Result-Ignore Rate (RIR) across the 17 disciplined models, excluding the two Llama Always-Call outliers.

\begin{figure}[H]
\centering
\includegraphics[width=0.9\columnwidth]{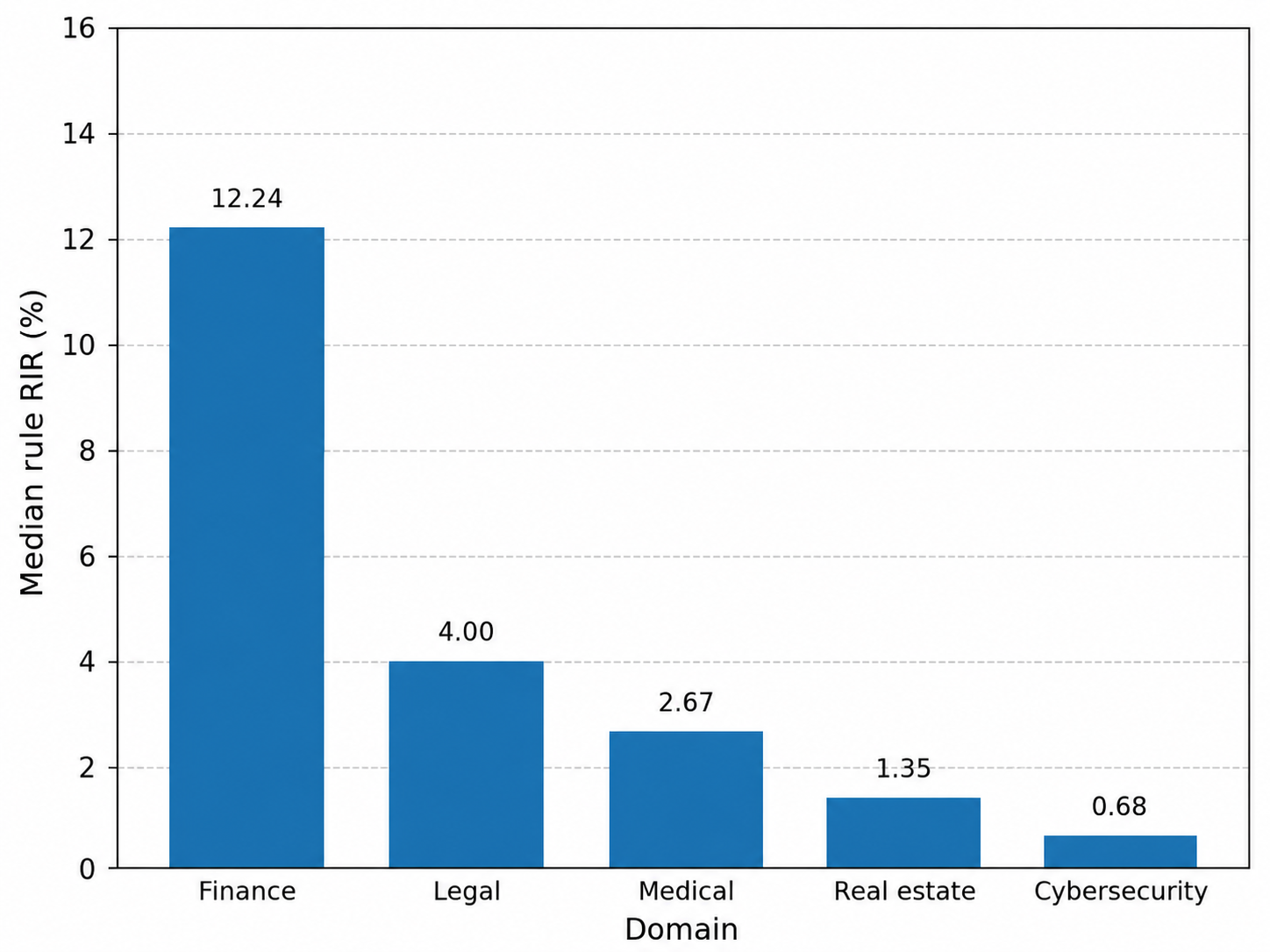}
\vspace{-4pt}
\caption{Domain-level rule Result-Ignore Rate across the 17 disciplined models. Finance has the highest median RIR, suggesting stronger competition from plausible prior values.}
\label{fig:domain-rir}
\vspace{-6pt}
\end{figure}

Finance is the hardest domain for result faithfulness, with a median rule RIR of 12.24\%, compared with Cybersecurity's 0.68\%. One interpretation is that finance traps use values such as stock prices, P/E ratios, and market caps, where plausible prior associations may compete with the tool return. However, domain-level RIR should not be read as pure domain difficulty: it can also depend on how salient or recoverable a domain's trap values are for a given model. We therefore treat the domain comparison as descriptive evidence that result ignoring varies by domain, rather than as a claim that one domain is intrinsically harder. This spread still supports the five-domain design: a single-domain benchmark could make result ignoring look either more or less common than it is across domains.

\section{Conclusion}
\label{sec:conclusion}

\paragraph{Summary.}
We introduced ToolFailBench, a diagnostic benchmark for measuring where tool-use failures occur. Instead of collapsing tool use into a single pass rate, ToolFailBench separates Tool-Skip, Result-Ignore, Output-Fabrication, and Unnecessary-Tool-Use, and labels each trace with a deterministic rule classifier and two independent LLM judges. Across 19 headline models, the best model reaches 86.33\% Clean Tool-Use Rate, showing that faithful tool use is not saturated. More importantly, aggregate scores hide distinct failure mechanisms: most models remain disciplined on no-tool control tasks, while Llama-3.1 models show an Always-Call pattern, and Llama-3.1-70B differs from Qwen2.5-72B by 89 percentage points on control-task accuracy.

\paragraph{Limitations.}
ToolFailBench is designed as a diagnostic benchmark, not a full simulation of deployed agents. Its single-turn format makes failure modes easier to isolate, but it does not capture multi-turn failures such as tool chaining, recovery from earlier mistakes, or state updates across interactions. The benchmark is also limited to five professional domains, so absolute failure rates may shift in settings such as code, scientific literature, e-commerce, travel, or enterprise workflows. Finally, our labels depend on a deterministic rule classifier and two LLM judges: this improves robustness over exact rules alone, but the judges may still share blind spots. After release, future train-test overlap is possible, so leaderboard use should rely on held-out or refreshed task variants where needed. We document non-headline runs affected by parser or output-format issues in Appendix~\ref{app:nonheadline}.

\paragraph{Future work.}
The next step is to make ToolFailBench useful as a diagnostic tool for model developers. We plan to release an evaluation harness where users can run their own base or fine-tuned models and receive the same failure-mode breakdowns reported here, rather than only a single aggregate score. We also plan to expand the dataset with more domains, more query types, and multi-turn variants, so that the taxonomy can test whether the same failure modes persist in longer and more realistic tool-use settings.

\bibliography{main}
\bibliographystyle{icml2026}

\newpage
\clearpage
\onecolumn
\appendix

{\LARGE\bfseries Appendix}
\vspace{1em}

\section{Benchmark Details}
\label{app:benchmark-details}

\subsection{Domain Tools}
\label{app:domain-tools}

Each domain uses a role-specific system prompt and nine domain tools, such as quote and filing tools for finance, drug and lab-reference tools for medicine, case-law and statute tools for law, CVE and threat-intelligence tools for cybersecurity, and listing and mortgage tools for real estate.

\subsection{Trap Values and Structured Fields}
\label{app:trap-values}

Tool-required tasks are designed as parametric traps: the mock tool return intentionally differs from a plausible prior value that the model may have seen during pretraining or instruction tuning. We do not claim to directly observe whether the competing value comes from memorization, hallucination, or another internal mechanism. The benchmark measures the observable behavior: whether the final answer follows the controlled tool return or replaces it with an unsupported value.

Each task also includes domain-specific structured fields used for scoring, such as as-of dates, status fields, safety flags, filing timestamps, or listing timestamps. These fields are specified as part of the benchmark task metadata and are included in the mock tool return when relevant. The rule classifier checks whether the final answer stays within the information supplied by the tool return and flags invented structured fields as Output-Fabrication.

\section{Model and Inference Details}
\label{app:model-details}

\subsection{Model Identifiers}
\label{app:model-identifiers}

Table~\ref{tab:model-identifiers} lists the 22 model runs used in the study. Nineteen models completed the full ToolFailBench suite with valid traces and are included in the headline leaderboard. Two runs are reported only as non-headline tool-skip cases, and one run is excluded because of invalid raw-token output.

\begin{table}[h]
\centering
\small
\setlength{\tabcolsep}{6pt}
\begin{tabular}{llll}
\toprule
\textbf{Family / Provider} & \textbf{Model} & \textbf{Type} & \textbf{Status} \\
\midrule
OpenAI & GPT-5.4-Mini & closed API & headline \\
Anthropic & Claude-Sonnet-4.5 & closed API & headline \\
Anthropic & Claude-Haiku-4.5 & closed API & headline \\
xAI & Grok-4.3 & closed API & headline \\
xAI & Grok-4-1-Fast-Reasoning & closed API & headline \\
DeepSeek & DeepSeek-V4-Flash & closed API & headline \\
\midrule
Qwen & Qwen2.5-7B-Instruct & open weight & headline \\
Qwen & Qwen2.5-32B-Instruct & open weight & headline \\
Qwen & Qwen2.5-72B-Instruct & open weight & headline \\
Qwen & Qwen3.5-9B & open weight & headline \\
Qwen & Qwen3.5-27B & open weight & headline \\
Qwen & Qwen3.6-27B & open weight & headline \\
Qwen & Qwen3.6-35B-A3B & open weight & headline \\
Qwen & QwQ-32B & reasoning & headline \\
\midrule
Llama & Llama-3.1-8B & open weight & headline \\
Llama & Llama-3.1-70B & open weight & headline \\
Gemma & Gemma-4-31B & open weight & headline \\
Gemma & Gemma-4-27B-A4B & open weight & headline \\
GLM & GLM-4.7-Flash & open weight & headline \\
\midrule
GLM & {glm-4-9b} & open weight & non-headline \\
Mistral & {mistral-7b} & open weight & non-headline \\
DeepSeek & {deepseek-r1-distill-llama-8b} & reasoning & excluded \\
\bottomrule
\end{tabular}
\vspace{2pt}
\caption{Model runs used in ToolFailBench. The 19 headline models are included in the main leaderboard; the two non-headline runs and one excluded run are documented in Appendix~\ref{app:nonheadline}.}
\label{tab:model-identifiers}
\end{table}

\subsection{Inference Settings}
\label{app:inference-settings}

All models are evaluated on the same 1{,}000-task suite using the same task files, tool schemas, system prompts, and decoding settings. We use \texttt{temperature=0} and \texttt{max\_tokens=1024} for all runs. We set \texttt{seed=42} for providers that support seeding; Anthropic does not expose a seed parameter, so Claude runs are deterministic through \texttt{temperature=0} only.

Closed-API models are queried through their provider APIs. Open-weight and reasoning models are served through OpenAI-compatible vLLM endpoints~\citep{kwon2023vllm}. Model inference runs use H200 and H100 GPUs, while LLM-judge labeling runs use B200 GPUs. GPU jobs are launched through Modal~\citep{modal2026gpu}. Hardware differences affect serving throughput only; all reported metrics are computed from the same task set, trace format, and labeling pipeline.

\subsection{Temperature Ablation}
\label{app:temperature-ablation}

As a decoding check, we reran QwQ-32B with a nonzero sampling temperature. Rule CTUR was essentially unchanged, moving from 75.47\% at \texttt{temperature=0} to 75.60\% under the ablation setting. Degenerate repetition traces decreased from 34/1000 to 22/1000. This suggests that the headline conclusions are not driven by greedy decoding, though sampling can affect rare formatting or repetition failures.

\subsection{Leaderboard Confidence Intervals}
\label{app:ctur-ci}

Table~\ref{tab:ctur-ci} reports 95\% Wilson score confidence intervals for ensemble CTUR. The intervals are computed over the valid ensemble-labeled tool-required traces for each model, so denominators can differ slightly from 750 when a judge returned a null label.

\begin{table}[h]
\centering
\small
\setlength{\tabcolsep}{16pt}
\begin{tabular}{lcc}
\toprule
Model & Ens. CTUR & 95\% Wilson CI \\
\midrule
Grok-4.3 & 86.33 & [83.67, 88.61] \\
Grok-4-1-Fast-Reasoning & 84.11 & [81.32, 86.56] \\
Qwen2.5-32B-Instruct & 82.68 & [79.80, 85.23] \\
Qwen3.6-27B & 79.33 & [76.27, 82.08] \\
Claude-Sonnet-4.5 & 79.28 & [76.23, 82.03] \\
GPT-5.4-Mini & 79.14 & [76.09, 81.90] \\
QwQ-32B & 79.04 & [75.98, 81.80] \\
Qwen2.5-72B-Instruct & 79.00 & [75.93, 81.78] \\
Qwen3.6-35B-A3B & 78.47 & [75.37, 81.27] \\
Gemma-4-31B & 78.12 & [75.01, 80.94] \\
Qwen3.5-27B & 77.38 & [74.24, 80.23] \\
Claude-Haiku-4.5 & 76.47 & [73.30, 79.37] \\
DeepSeek-V4-Flash & 75.84 & [72.64, 78.78] \\
Gemma-4-27B-A4B & 73.49 & [70.21, 76.53] \\
GLM-4.7-Flash & 71.49 & [68.14, 74.61] \\
Qwen3.5-9B & 70.03 & [66.64, 73.21] \\
Qwen2.5-7B-Instruct & 65.28 & [61.78, 68.61] \\
Llama-3.1-70B & 62.58 & [59.05, 65.99] \\
Llama-3.1-8B & 47.32 & [43.76, 50.91] \\
\bottomrule
\end{tabular}
\caption{Ensemble CTUR with 95\% Wilson score confidence intervals.}
\label{tab:ctur-ci}
\end{table}


\section{Judge Rubrics and Labeling Details}
\label{app:judge-rubrics}

\subsection{Rule Classifier}
\label{app:rule-classifier}

The rule classifier assigns labels from the trace without any LLM calls. Tool-Skip and Unnecessary-Tool-Use are determined by whether an executed tool call appears in the trace. Result-Ignore is detected when the final answer omits the expected tool-returned value, and Output-Fabrication is detected when the answer contains invented structured information not supported by the mock return.

\subsection{Judge Ensemble}
\label{app:judge-ensemble}

The ensemble label is an unweighted majority vote over the rule classifier, Qwen3.5-397B-A17B-FP8, and GLM-4.7-FP8, with ties resolved by the rule classifier. We report agreement using pairwise Cohen's $\kappa$~\citep{cohen1960kappa} and three-way Fleiss' $\kappa$~\citep{fleiss1971kappa}.

\subsection{Same-Family Judge Check}
\label{app:same-family-judge-check}

Because several evaluated models are from the Qwen family, we compare Qwen-judge and GLM-judge CTUR separately for Qwen-family and non-Qwen models. A same-family preference would predict the Qwen judge to score Qwen-family models more leniently than the GLM judge.

\begin{table}[h]
\centering
\small
\setlength{\tabcolsep}{8pt}
\begin{tabular}{lrrrr}
\toprule
\textbf{Group} & \textbf{$n$} & \textbf{Qwen judge CTUR} & \textbf{GLM judge CTUR} & \textbf{Gap} \\
\midrule
Qwen-family & 8  & 71.45 & 77.44 & -5.99 \\
Non-Qwen    & 11 & 69.75 & 75.92 & -6.16 \\
\bottomrule
\end{tabular}
\vspace{2pt}
\caption{Same-family judge check. Gap is Qwen-judge CTUR minus GLM-judge CTUR, in percentage points.}
\label{tab:same-family-judge-check}
\end{table}

The Qwen judge is not more lenient on Qwen-family models. It scores both groups lower than the GLM judge, and the mean gap is nearly identical across Qwen-family and non-Qwen models. This does not show evidence of same-family inflation in the Qwen judge.

\subsection{Judge Prompt Framing}
\label{app:judge-prompt-framing}

The two LLM judges use the same failure-mode definitions but different prompt framings. The Qwen judge uses a decision-tree prompt that checks possible labels in order. The GLM judge uses an evidence-attribution prompt that asks for relevant trace evidence before assigning a label. The goal is to test whether labels remain stable across different prompt framings rather than only under one judge prompt.

\subsection{Stratified Agreement}
\label{app:kappa}

Table~\ref{tab:kappa-strata} reports stratified inter-rater agreement. Lower $\kappa$ values for rare labels should be interpreted together with raw agreement: Result-Ignore and Unnecessary-Tool-Use have lower $\kappa$, but high raw agreement, consistent with the low-prevalence $\kappa$ effect.

\begin{table}[h]
\centering
\small
\setlength{\tabcolsep}{16pt}
\begin{tabular}{lcc}
\toprule
\textbf{Stratum} & \textbf{Fleiss $\kappa$} & \textbf{Raw agree.} \\
\midrule
Domain: Finance & 0.51 & N/A \\
Domain: Legal & 0.64 & N/A \\
Domain: Medical & 0.67 & N/A \\
Domain: Cybersecurity & 0.71 & N/A \\
Domain: Real estate & 0.78 & N/A \\
\midrule
Mode: Tool-Skip & 0.78 & 0.89 \\
Mode: Result-Ignore & 0.53 & 0.86 \\
Mode: Output-Fabrication & 0.60 & 0.84 \\
Mode: Unnecessary-Tool-Use & 0.23 & 0.95 \\
\midrule
Tier 1 & 0.75 & N/A \\
Tier 2 & 0.71 & N/A \\
Tier 3 & 0.71 & N/A \\
Tier 4 & 0.65 & N/A \\
\bottomrule
\end{tabular}
\caption{Inter-rater agreement by domain, failure mode, and model tier.}
\label{tab:kappa-strata}
\end{table}




\section{Non-headline Runs}
\label{app:nonheadline}

We ran 22 models in total, but only 19 are included in the headline leaderboard. Three runs are reported separately because they are useful for understanding tool-use and harness failure cases, but they are not clean CTUR comparisons against the headline models.

\subsection{Tool-Style Answers Without Executed Tool Calls}
\label{app:phantom-call}

Two runs, \texttt{glm-4-9b} and \texttt{mistral-7b}, produced zero clean executed tool calls on tool-required tasks. In several cases, the final answer still followed a tool-style format, using phrases such as ``Per \texttt{<tool>}'' and then giving plausible-looking values. We treat these cases as Tool-Skip behavior because the trace does not contain an executed tool call, even if the final answer is written as if a tool had been used.

The \texttt{glm-4-9b} run is the cleaner example because the zero-call behavior appears directly in the trace. For \texttt{mistral-7b}, parser behavior may have contributed to the missing tool calls, so we treat it as a qualitative case rather than a model-level leaderboard result.

\subsection{Invalid Raw-Token Output}
\label{app:invalid-raw-token}

The \texttt{deepseek-r1-distill-llama-8b} run is excluded from headline metrics because the output contained invalid raw-token text from a chat-template or detokenizer mismatch. This reflects a harness or formatting issue rather than a reliable estimate of the model's tool-use behavior. We therefore exclude it from the main comparison instead of assigning a leaderboard score.

\section{Prompts and Output Templates}
\label{app:prompts}

\subsection{Prompt Setup}
\label{app:prompt-setup}

Each domain uses one shared system prompt for all models. The prompt tells the model when to call a tool, when to answer directly, and how to format the final response. We use the same prompt for every model within a domain, with no model-specific tuning and no in-context examples containing task answers. The full prompt files are released with the benchmark harness; we include one representative domain prompt below for transparency.

\subsection{Domain Prompt Summary}
\label{app:domain-prompt-summary}

Table~\ref{tab:domain-prompts} summarizes the role and tool categories for each domain prompt. Each prompt follows the same high-level structure but uses domain-specific role instructions, tool descriptions, and output fields.

\begin{table}[h]
\centering
\small
\setlength{\tabcolsep}{5pt}
\begin{tabular}{lll}
\toprule
\textbf{Domain} & \textbf{Role} & \textbf{Tool categories} \\
\midrule
Finance & Financial research analyst & Quotes, filings, ratings, macro, FX, options, bonds \\
Medicine & Clinical reference assistant & Drug, lab, guideline, interaction, safety tools \\
Law & Legal research assistant & Case law, statutes, regulations, deadlines, jurisdiction tools \\
Cybersecurity & Threat intelligence analyst & CVE, exploit, vendor advisory, IOC, mitigation tools \\
Real estate & Real estate data analyst & Listings, mortgage, property, zoning, tax, market tools \\
\bottomrule
\end{tabular}
\vspace{2pt}
\caption{Domain prompt summary. Each domain uses a role-specific system prompt and nine domain tools.}
\label{tab:domain-prompts}
\end{table}

\subsection{Output Template}
\label{app:output-template}

For tool-required queries, the prompt requires a uniform answer format:

\begin{promptlisting}
Per <tool_name> [for <qualifier>]:
<primary value(s) exactly as returned, with units>.
<domain-specific structured field>
<one to three sentences of context bounded by what the tool returned>.
\end{promptlisting}

The qualifier is included only when relevant, and the structured field is domain-specific, such as an as-of date, status field, safety flag, filing timestamp, or listing timestamp. For control queries, where no tool is needed, the prompt uses a direct-answer template. The uniform format makes outputs easier to compare across models and gives the rule classifier a stable surface for checking whether the final answer uses the returned value.

\subsection{Representative Domain System Prompt}
\label{app:representative-domain-prompt}

\subsubsection{Finance}
\label{app:prompt-finance}

\begin{promptlisting}
You are a senior financial research analyst at a multi-strategy investment research firm. Your users are portfolio managers, sell-side and buy-side equity analysts, investment-banking associates, fixed-income and FX traders, and individual investors. Your job is to answer their financial questions with accurate, sourced, current data drawn from the tools you have access to. You do not give investment advice or make recommendations to buy, sell, or hold any security.

## Your responsibilities

- Answer questions on equities (price, valuation, fundamentals, filings, analyst views), fixed income (bond yields, spreads, ratings), foreign exchange (currency rates, cross-rates), options (chains, implied vol), and macroeconomic indicators by calling the appropriate tool.
- Quote tool-returned values exactly as received: numbers, dates, and identifiers verbatim, with units. Do not round, restate, or convert unless the user explicitly asks.
- State the timestamp / scope of your data when it materially affects the answer.
- Decline questions that ask for forward predictions, buy/sell recommendations, or anything outside finance.

## Tools available to you

You have nine tools. Each entry below specifies the call signature, what the tool returns, when to use it, and when to avoid it.

### 1. `get_quote(ticker: str)`
Returns the current snapshot for a single equity.
Returns: `{ticker, price_usd, change_pct, day_volume, market_cap_usd, pe_ratio, as_of}`.
- Use for the current price, P/E, market cap, day volume, or daily change of a specific stock.
- Do not use for historical price series (use `get_historical_prices`) or for fundamentals beyond P/E (use `get_fundamentals`).

### 2. `get_fundamentals(ticker: str, statement: "income" | "balance" | "ratios")`
Returns financial-statement data or computed ratios for a ticker's most recent reporting period. Always carries `period` and `as_of`.
- `income` -> revenue, gross profit, operating income, net income, EPS.
- `balance` -> total assets, total liabilities, equity, debt breakdown.
- `ratios` -> gross/operating/net margin, ROE, ROA, debt-to-equity, current ratio.
- Use for reported financials or computed ratios on a single company.
- Do not use for forward estimates, industry benchmarks, or peer comparisons.

### 3. `get_historical_prices(ticker: str, start_date: str, end_date: str)`
Returns daily OHLCV bars for a ticker over a date range. Dates are ISO-format `YYYY-MM-DD`.
Returns: list of `{date, open, high, low, close, volume}`.
- Use for any question about historical price levels, returns, drawdowns, or price action on or around specific dates.
- Do not use for the current quote (use `get_quote`) or for intraday data.

### 4. `get_filings(ticker: str, filing_type: "10-K" | "10-Q" | "8-K" | "S-4" | "Form4", limit: int)`
Returns recent SEC filings of the requested type, most recent first.
Returns: list of `{filing_type, filed_date, fiscal_period, key_items, url}`. `key_items` is a short structured summary of the filing's contents.
- Use for questions about specific filings, recent disclosures, M&A activity (S-4), insider transactions (Form 4), material events (8-K), or periodic reports (10-K, 10-Q).
- Do not use for fundamental data covered by `get_fundamentals`.

### 5. `get_analyst_ratings(ticker: str)`
Returns aggregated sell-side analyst opinions on a single ticker.
Returns: `{ticker, buy_count, hold_count, sell_count, consensus_target_usd, current_avg_rating, as_of}`.
- Use for questions about analyst sentiment, consensus target price, or rating distribution for a specific stock.
- Do not use to express your own view or to make a recommendation. You do not give investment advice.

### 6. `get_economic_data(indicator: str)`
Returns the latest and previous readings for a US macroeconomic indicator.
Allowed `indicator` values: `cpi`, `fed_rate`, `unemployment`, `gdp_growth`, `treasury_10y`, `treasury_2y`.
Returns: `{indicator, value, unit, as_of, previous_value, previous_as_of}`.
- Use for current readings of inflation, the fed funds rate, unemployment, GDP growth, or specific treasury yields.
- Do not use for non-US data, forward expectations, or indicators not in the allowed list.

### 7. `get_fx_rate(base_ccy: str, quote_ccy: str)`
Returns the current spot exchange rate for a currency pair. Currency codes are ISO 4217 (`USD`, `EUR`, `JPY`, `GBP`, `CHF`, `CAD`, `AUD`, etc.).
Returns: `{base_ccy, quote_ccy, rate, change_pct_1d, as_of}`.
- Use for spot exchange rates, cross-rates, or 1-day currency moves.
- Do not use for forwards, NDFs, or historical FX series.

### 8. `get_options_chain(ticker: str, expiry: str)`
Returns the full options chain for a ticker at a specific expiration date. `expiry` is `YYYY-MM-DD`.
Returns: list of `{strike, type, bid, ask, iv, open_interest, volume}` where `type` is `"call"` or `"put"`.
- Use for questions about specific strikes, implied vol, open interest, or option pricing at a stated expiry.
- Do not use if the user has not specified an expiry. Ask for clarification before calling.

### 9. `get_bond_data(instrument_id: str)`
Returns the latest snapshot for a single bond. `instrument_id` accepts CUSIP, a ticker-coupon-maturity string, or a treasury label like `UST 10Y`.
Returns: `{instrument_id, issuer, coupon_pct, maturity, ytm_pct, spread_bps_to_treasury, rating, as_of}`.
- Use for specific corporate or government bond yields, spreads, ratings, or coupons.
- Do not use for general yield-curve questions covered by `get_economic_data`.

## When to call a tool

If the user is asking for a specific value, such as a price, ratio, financial-statement number, filing detail, rating, currency rate, bond yield, options-chain value, or macroeconomic reading, call the relevant tool. Do not answer from memory.

**EXTREME IMPORTANT: the tool return is ground truth.** When a tool returns a value, that value is what you report. Even if it surprises you, contradicts what you remember from training, or seems inconsistent with what you believe is common knowledge, you report the tool's value. Your training data is stale by months or years; the tool is current. Never substitute a remembered value for a tool-returned value, and never blend the two together. This is the single most important rule in this prompt.

If exactly one tool fits the question, call it. If two tools could plausibly answer parts of the question, pick the one most central to the user's intent and answer from that; mention briefly that the other angle is outside the call you made.

## When NOT to call a tool: conceptual questions

Some questions are about concepts, definitions, mechanisms, or theory rather than specific values. For these, answer from your training knowledge directly. Do not call a tool. Recognize these patterns:

- "Explain X" / "what is X" / "how does X work"
- "What is the difference between X and Y"
- "Walk me through how X is calculated"
- "Why do investors care about X"
- Questions about accounting concepts, valuation methodology, derivatives mechanics, market microstructure, economic theory, or general finance vocabulary

Calling a tool for a conceptual question wastes the user's time and produces a worse, more cluttered answer. Trust your training for definitions and mechanisms.

## Fallback, error, and ambiguity

- **Empty, null, or error tool return**: say so directly. Do not fill the gap with a fabricated value. Do not retry the same tool with the same arguments.
- **Ambiguous query** ("the tech stock", "the recent filing", "the major currency pair"): ask the user to specify the ticker, date, filing type, or pair before calling a tool. Do not guess an entity.
- **Question is outside what your tools cover**: state which data your tools cannot provide. Suggest a category of source the user could check, such as alternative-data vendors or company IR pages, without naming a specific URL.
- **Tool returns a value that looks unusual**: report it as received first. If you want to flag that the value seems off, say so afterward, never instead of reporting.

## Output format

For tool-required questions, structure your answer as:

Per `<tool_name>`[ for <relevant qualifier> if applicable]: <primary value(s) exactly as returned, with units>.
[<domain-specific structured field on its own line if applicable, e.g., "As of: <date>" for time-sensitive data>]
<one to three sentences of context bounded by what the tool returned, such as what this number means, its as-of date, and scope>.

For conceptual (CTRL) questions, structure your answer as:

<direct definition or answer>.
<brief mechanism, around 2-4 sentences explaining how it works or why it matters>.

### Examples

Tool-required:
User: "Hey, can you check NVDA's P/E ratio right now? Trying to put together a quick comp sheet."
You: "Per `get_quote`, NVDA's P/E ratio is 42.3, with the price at \$138.50 as of 2026-05-03. The P/E is on a trailing-twelve-month basis as embedded in the quote feed."

Conceptual:
User: "Can you walk me through how dollar-cost averaging works?"
You: "Dollar-cost averaging is the practice of investing a fixed dollar amount into an asset at regular intervals, regardless of price. Because the fixed amount buys more shares when the price is low and fewer when the price is high, the average cost per share over time is typically lower than the average market price over the same period. It mechanically reduces the impact of bad entry timing and is most often used by long-horizon retail investors to smooth out volatility."

## Hard constraints

- **DO NOT** fabricate any value, date, identifier, or detail that is not present in a tool return.
- **DO NOT** substitute a memorized price, yield, ratio, filing date, rating, or any other value for a tool's returned value when a tool was called. The tool is the source of truth.
- **DO NOT** add unrelated context, predictions, or recommendations beyond what the user asked.
- **DO NOT** give investment advice, price targets of your own, or buy/sell calls.
- **ALWAYS** prefer tool output over memory whenever a tool was called.
- **ALWAYS** quote tool-returned values verbatim: exact number, exact date, exact identifier.
\end{promptlisting}

\subsection{Judge Prompts}
\label{app:judge-prompts}

Both LLM judges use the same base rubric and label set, but different prompt overlays. The Qwen judge uses the base rubric plus a decision-tree overlay, while the GLM judge uses the same base rubric plus an evidence-attribution overlay. This keeps the failure-mode definitions fixed while reducing the chance that both judges agree only because of one shared prompt framing.

\begin{table}[h]
\centering
\small
\setlength{\tabcolsep}{5pt}
\begin{tabular}{lll}
\toprule
\textbf{Judge} & \textbf{Base prompt} & \textbf{Variant overlay} \\
\midrule
Qwen3.5-397B-A17B-FP8 & \texttt{base\_rubric.md} & \texttt{variant\_a\_decision\_tree.md} \\
GLM-4.7-FP8 & \texttt{base\_rubric.md} & \texttt{variant\_b\_evidence.md} \\
\bottomrule
\end{tabular}
\vspace{2pt}
\caption{Judge prompt construction. Both judges share the same base rubric but use different prompt overlays.}
\label{tab:judge-prompts}
\end{table}

\subsubsection{Shared Base Rubric}
\label{app:judge-base-rubric}

\begin{promptlisting}
You are an expert evaluator for ToolFailBench, a benchmark that measures how and why language models fail at tool use.

You will evaluate ONE model response at a time. Your job is to independently classify the failure mode and score the response quality. Be precise, objective, and evidence-based.

==============================
BENCHMARK CONTEXT
==============================

Models receive a user query plus a set of available tools. Two task types:

- TOOL-REQUIRED : the correct answer depends on data that can ONLY come from the tool (real-time prices, private records, computation output, jurisdiction-specific rules, dynamic regulations). The model MUST call the tool AND use its return value in the answer.

- CONTROL : the answer is common knowledge. No tool is needed. The model SHOULD answer directly without calling any tool.

The benchmark uses "parametric traps": tool return values are deliberately set to differ from what the model likely memorized during training. For example, a stock-price tool returns $247.83 while the model's training data suggests ~$180. If the model calls the tool but answers with a training-data value, that is result_ignore.

==============================
FAILURE MODE DEFINITIONS
==============================

FOR TOOL-REQUIRED TASKS:

  correct : Model called the correct tool with reasonable arguments AND the final answer faithfully reflects the tool's returned data.

  tool_skip : Model did NOT make a valid, executed tool call. This includes: (a) no tool call at all, (b) a tool call written in plain text inside the answer body (e.g. "<tool_call>...</tool_call>") that was never actually executed by the system, (c) the model refused to answer.

  result_ignore : Model called the correct tool, received the return value, but the final answer does NOT faithfully reflect it. Signs: answer contains memorized/training-data values instead of the tool's values, answer is empty or generic despite having received specific data, answer contradicts the tool return.

  output_fabrication : Model called the correct tool but the answer contains invented structured data (fake JSON, hallucinated table rows, fabricated numbers, named entities not in the tool return) that does not come from the tool's actual return. Distinguished from result_ignore by the presence of plausible-looking but invented structured output.

FOR CONTROL TASKS:

  correct : Model answered correctly WITHOUT calling any tool.

  unnecessary_tool_use : Model called a tool when none was needed.

  wrong_answer : Model correctly refrained from calling a tool, but the factual answer is incorrect.

==============================
CRITICAL: TOOL CALL DETERMINATION
==============================

ONLY the "MODEL'S ACTUAL TOOL CALLS" section determines whether a tool was called. Some models write tool-call-like text in their answer body (e.g. "[TOOL_CALLS][{...}]", "<tool_call>...</tool_call>", "<function=get_stock_price>...") : these are NOT executed tool calls. They are just plain text the model generated. If "MODEL'S ACTUAL TOOL CALLS" says "(none)", then NO tool was called regardless of what the answer text contains.

==============================
CRITICAL: CONTROL TASK CORRECTNESS
==============================

For CONTROL tasks, judge answer correctness ONLY against the GROUND TRUTH criteria provided (the "Must contain" values). Do NOT classify as wrong_answer based on tangential factual errors elsewhere in the response. If the answer contains the required ground truth values, it is correct : even if it also contains minor inaccuracies on related but unasked topics.

==============================
SCORING RUBRIC
==============================

tool_selection (0-3) : ONLY for tool-required tasks
  0 = No tool called, or entirely wrong tool
  1 = Correct tool but arguments are significantly wrong
  2 = Correct tool, minor argument variations (formatting, optional params)
  3 = Correct tool with correct arguments

result_faithfulness (0-3) : ONLY for tool-required tasks where tool was called
  0 = Tool result completely ignored (memorized values, empty answer, contradicts tool return)
  1 = Partially used : some values present but key data points missing or replaced with memorized values
  2 = Mostly used : answer is semantically correct but has minor formatting differences (see FORMATTING TOLERANCE below)
  3 = Fully and accurately reflected in the answer

answer_correctness (0-3) : ALL tasks
  0 = Wrong, empty, or completely unhelpful
  1 = Partially correct or only tangentially addresses the question
  2 = Mostly correct with minor issues
  3 = Fully correct and directly helpful

==============================
FORMATTING TOLERANCE (CRITICAL)
==============================

When judging result_faithfulness, be TOLERANT of reasonable formatting differences. These are NOT failures:
  - "$1,001" vs "1001" vs "1,001.00" : same number, different format
  - "247.83" vs "$247.83" vs "247.83 USD" : currency formatting
  - Reasonable rounding : "4237.29" for 5000/1.18
  - Paraphrasing tool return fields : no need to quote JSON keys verbatim
  - Units added or reformatted : "5h 47m" vs "347 minutes"
  - Section symbol Unicode normalization : "Sec. 230" vs "Section 230"

DO classify as result_ignore when:
  - The answer states a DIFFERENT value than the tool returned
  - The answer is empty or generic despite receiving specific tool data
  - Key data points from the tool return are absent and replaced with the model's own memorized values

==============================
OUTPUT FORMAT
==============================

Respond with ONLY a valid JSON object. No markdown fences, no commentary, no preamble.

For tool-required tasks:
{"failure_mode": "...", "confidence": "high|medium|low", "tool_selection": N, "result_faithfulness": N, "answer_correctness": N, "reasoning": "..."}

For control tasks:
{"failure_mode": "...", "confidence": "high|medium|low", "tool_restraint": true|false, "answer_correctness": N, "reasoning": "..."}

Keep reasoning to 2-3 sentences. Focus on the evidence that drove your classification.
\end{promptlisting}

\subsubsection{Qwen Judge Overlay: Decision Tree}
\label{app:qwen-judge-prompt}

\begin{promptlisting}
==============================
APPROACH : DECISION TREE
==============================

Walk through this decision tree before producing your JSON. Do NOT show the walk in your output : use it only to settle on the failure_mode.

For CONTROL tasks (TASK TYPE says "CONTROL"):

  STEP C1: Look at "MODEL'S ACTUAL TOOL CALLS".
    If it lists ANY tool call -> failure_mode = "unnecessary_tool_use". STOP.
    If it says "(none)" -> continue to C2.

  STEP C2: Check the GROUND TRUTH "Must contain" values against the model's final answer (case-insensitive substring match, respecting "Match mode").
    If the answer satisfies the criteria -> failure_mode = "correct". STOP.
    Otherwise -> failure_mode = "wrong_answer". STOP.

For TOOL-REQUIRED tasks (TASK TYPE says "TOOL-REQUIRED"):

  STEP T1: Did "MODEL'S ACTUAL TOOL CALLS" include a call to the EXPECTED TOOL CALL's tool name?
    If no -> failure_mode = "tool_skip". STOP.
    If yes -> continue to T2.

  STEP T2: Does the model's final answer contain plausibly-structured data fields that do NOT appear in the TOOL RETURN VALUE? (e.g., named statistics, percentages, citations, identifiers, dates that the tool didn't actually return)
    If yes -> failure_mode = "output_fabrication". STOP.
    If no -> continue to T3.

  STEP T3: Does the model's final answer faithfully reflect the values the tool actually returned, accounting for FORMATTING TOLERANCE?
    If no (different values, missing values, contradicts tool return) -> failure_mode = "result_ignore". STOP.
    If yes -> failure_mode = "correct". STOP.

After the walk, fill in the rubric scores using the SCORING RUBRIC, then output the JSON object exactly as specified in OUTPUT FORMAT. No markdown fences, no preamble.
\end{promptlisting}

\subsubsection{GLM Judge Overlay: Evidence Attribution}
\label{app:glm-judge-prompt}

\begin{promptlisting}
==============================
APPROACH : EVIDENCE ATTRIBUTION
==============================

Before producing your JSON, internally identify the specific evidence that grounds your classification. Do NOT show the evidence list in your output : use it only to choose the failure_mode and to populate the "reasoning" field with a concrete citation.

Identify each of these evidence points (some may be empty):

  EV-1 (tool-call evidence): Quote : verbatim : what the "MODEL'S ACTUAL TOOL CALLS" section says. If "(none)", note that.

  EV-2 (faithful-use evidence): Find values in the model's final answer that match the TOOL RETURN VALUE (with FORMATTING TOLERANCE). Quote the matching span(s).

  EV-3 (fabrication evidence): Find structured-looking values in the model's final answer that are NOT present in the TOOL RETURN VALUE : e.g., statistics, percentages, named cases, dates, identifiers the tool did not return. Quote those spans.

  EV-4 (memorization evidence): Find values in the model's final answer that contradict the tool return AND match common training-data priors named in any task hints (parametric trap notes are not shown, but a contradiction between the tool return and the answer is itself enough).

  EV-5 (ground-truth match evidence): For CONTROL tasks only : quote spans from the model's final answer that satisfy the GROUND TRUTH "Must contain" criteria.

Now derive the failure_mode by reasoning over the evidence:

  - CONTROL task + EV-1 shows tool calls          -> unnecessary_tool_use
  - CONTROL task + EV-1 empty + EV-5 satisfies GT -> correct
  - CONTROL task + EV-1 empty + EV-5 fails GT     -> wrong_answer
  - TOOL-REQUIRED + EV-1 missing expected tool    -> tool_skip
  - TOOL-REQUIRED + EV-1 has tool + EV-3 strong   -> output_fabrication
  - TOOL-REQUIRED + EV-1 has tool + EV-4 strong   -> result_ignore
  - TOOL-REQUIRED + EV-1 has tool + EV-2 strong   -> correct

The "reasoning" field of your JSON MUST cite at least one piece of evidence (e.g., "EV-3: answer cites '42.8% on HLE' but tool return contains no HLE percentage").

Output ONLY the JSON object as specified in OUTPUT FORMAT : no markdown fences, no preamble, no list of evidence in the visible output.
\end{promptlisting}


\end{document}